\documentclass[a4paper,11pt]{article}
\usepackage{ijcnlp2013}
\usepackage{graphicx}
\usepackage{epstopdf}
\usepackage{times}
\usepackage{url}
\usepackage{latexsym}

\title{Measuring the Effect of Discourse Relations on Blog Summarization}

\author{Shamima Mithun \\
       Concordia University\\
      Montreal, Quebec, Canada\\
{\tt shamima.mithun@gmail.com}
\And
Leila Kosseim \\
       Concordia University\\
             Montreal, Quebec, Canada\\
{\tt kosseim@encs.concordia.ca}}

\date{}

\begin{document}
\maketitle
\begin{abstract}
The work presented in this paper attempts to evaluate and quantify the use of discourse relations in the context of blog summarization and compare their use to more traditional and factual texts. Specifically, we measured the usefulness of 6 discourse relations - namely \emph{comparison}, \emph{contingency}, \emph{illustration}, \emph{attribution}, \emph{topic-opinion}, and \emph{attributive} for the task of text summarization from blogs. We have evaluated the effect of each relation using the TAC 2008 opinion summarization dataset and compared them with the results with the  DUC 2007 dataset. The results show that in both textual genres, \emph{contingency}, \emph{comparison}, and \emph{illustration} relations provide a significant improvement on summarization content; while \emph{attribution}, \emph{topic-opinion}, and \emph{attributive} relations do not provide a consistent and significant improvement.
These results indicate that, at least for summarization, discourse relations are just as useful for informal and affective texts as for more traditional news articles.
\end{abstract}

\section{Introduction}
%For example, in the sentence
%``\emph{Where some operations in iPhoto take a few clicks in unexpected
%places, in Picasa they are almost always conveniently close to
%where you are currently working}.'' a \emph{contrast} relation is
%expressed.
It is widely accepted that in a coherent text, units should not be understood in isolation but in relation with each other through discourse relations that may or may not be explicitly marked. A text is not a linear combination of textual units but a hierarchically organized group of units placed together based on informational and intentional relations to one another. According to~\cite{Taboada06A}, ``Discourse relations - relations that
hold together different parts (i.e. proposition, sentence, or
paragraph) of the discourse - are partly responsible for the
perceived coherence of a text''.
For example, in the sentence \emph{``If you want the full Vista experience, you'll want a heavy system and graphics hardware, and lots of memory''}, the first and second clauses do not bear much meaning independently; but become more meaningful when we realize that they are related through the discourse relation \emph{condition}.

%To describe discourse relations, different theories have been developed such as Rhetorical Predicates~\cite{Grimes75}, Rhetorical Structure Theory~\cite{Mann88} and others (e.g. \cite{Hovy93}). Some theories are more inclusive than others
%with respect to the inventory and the definition of discourse structures and their applicability.
%For example, Rhetorical Structure Theory (RST)~\cite{Mann88} is more
%comprehensive compared to its predecessors because it provides
%extensive definitions of discourse relations and provides a computational approach to use these relations through the use of plans.
%However, the set of discourse relations proposed by
%these theories are often comparable.

Discourse relations have been found useful in many NLP applications such as natural language generation (e.g. \cite{McKeown85}) and news summarization (e.g.~\cite{Goldensohn06,Bosma04}) to improve coherence and better simulate human writing.
However, most of these work have been developed for formal, well-written and factual documents. Text available in the social media are typically written in a more casual style, are opinionated and speculative \cite{Andreevskaia07}. Because of this, techniques developed for formal texts, such as news articles, often do not behave as well when dealing with informal documents. In particular, news articles are more uniform in style and structure; whereas blogs often do not exhibit a
stereotypical discourse structure. As a result, for blogs, it is usually more difficult to identify and rank relevant units for summarization compared to news articles.

Several work have shown that discourse relations can improve the results of summarization in the case of factual texts or news articles (e.g. \cite{Otterbacher02}). However, to our knowledge no work has evaluated the usefulness of discourse relations for the summarization of informal and opinionated texts, as those found in the social media. In this paper, we consider the most frequent discourse relations found in blogs: namely \emph{comparison}, \emph{contingency}, \emph{illustration},
\emph{attribution}, \emph{topic-opinion}, and \emph{attributive} and evaluate the effect of each relation on informal text summarization using the Text Analysis Conference (TAC) 2008 opinion summarization dataset\footnote{http://www.nist.gov/tac/}. We then compare these results to those found with the news articles of the Document Understanding Conference (DUC) 2007 Main task dataset\footnote{http://www-nlpir.nist.gov/projects/duc/guidelines/2007.html}. The results show that in both types of texts, discourse relations seem to be as useful:  \emph{contingency}, \emph{comparison}, and \emph{illustration} relations provide a statistically significant improvement on the summary content; while the \emph{attribution}, \emph{topic-opinion}, and \emph{attributive} relations do not provide a consistent and significant improvement. 
\vspace{-2mm}

\section{Related Work on Discourse Relations for Summarization}
The use of discourse relations for text summarization is not new. Most notably,~\cite{Marcu97} used discourse relations for single document summarization and proposed
a discourse relation identification parsing algorithm. In some work (e.g.~\cite{Bosma04,Goldensohn06}),
discourse relations have been exploited successfully for multi-document summarization. In particular,~\cite{Otterbacher02} experimentally showed that discourse relations can improve the coherence of multi-document summaries. \cite{Bosma04} showed how discourse relations can be used effectively to incorporate additional contextual information
for a given question in a query-based summarization. \cite{Goldensohn06} used discourse relations for content
selection and organization of automatic summaries and achieved an improvement in both cases. Discourse relations were also used successfully by~\cite{Zahri11} for news summarization. 

However, the work described above have been developed for formal, well-written and factual documents. Most of these work show how discourse relations can be used in text summarization and show their overall usefulness. To the best of our knowledge, our work is the first to measure the effect of specific relations on the summarization of informal and opinionated text.

\section{Tagging Discourse Relations}\label{tag}
To evaluate the effect of discourse relations on a large scale, sentences need to be tagged automatically with discourse relations. For example, the sentence ``\emph{Yesterday, I stayed at home because it was raining}.''
needs to be tagged as containing a \emph{cause} relation. One sentence can convey zero or several discourse relations. For example, the sentence ``\emph{Starbucks has contributed to the popularity of good tasting coffee}'' does not contain any discourse relations of interest to us.
%\vspace{-4mm}
%begin{figure}[h]
%centering
%caption{A Sample RST Tree}
%~\\
%label{rrZillow}
%includegraphics[height=45mm]{figure/RSTZillow.eps}
%\includegraphics[height=65mm]{figure/RPF}
%\epsfig{file=figure/fig2}
%end{figure}
%vspace{-5mm}
%
On the other hand, the sentence ``\emph{While I like the Zillow interface and agree it's an easy way to find data, I'd prefer my readers used their own brain to perform a basic valuation of a property instead of relying on zestimates.}'' contains 5 relations of interest: one \emph{comparison}, three \emph{illustrations}, and one \emph{attribution}.
%This is illustrated in Figure \ref{rrZillow} which shows the RST tree for this sentence.

\subsection{Most Frequent Discourse Relations}
Since our work is performed within the framework of blog summarization; we have only considered the
discourse relations that are most useful to this application.
To find the set of the relations needed for this task, we have first manually analyzed 50 summaries randomly selected from participating systems at the
TAC 2008 opinion summarization track and 50 randomly selected blogs from BLOG06 corpus\footnote{http://ir.dcs.gla.ac.uk/test\_collections/blog06info.html}. In building
our relation taxonomy, we considered all main discourse relations
listed in the taxonomy of Mann and Thompson's Rhetorical Structure Theory (RST)~\cite{Mann88}. These discourse relations are also
considered in Grimes'~\cite{Grimes75} and Williams' predicate lists. From our corpus analysis, we have identified the six most prevalent discourse relations in this blog dataset,
namely \emph{comparison}, \emph{contingency}, \emph{illustration},
\emph{attribution}, \emph{topic-opinion}, and \emph{attributive}.
The \emph{comparison}, \emph{contingency}, and \emph{illustration} relations are also considered by most of
the work in the field of discourse analysis such as the PDTB: Penn Discourse TreeBank research group
\cite{Prasad08} and the RST Discourse Treebank research group~\cite{Carlson01}. We considered three additional
classes of relations: \emph{attributive}, \emph{attribution}, and
\emph{topic-opinion}. These discourse relations are summarized in Figure \ref{rr} while a description of these relations is given below.

%\vspace{-1mm}
\begin{figure}[h]
\centering
\caption{Most Frequent Discourse Relations in Blogs and their Sub-relations}
%~\\
\label{rr}
\includegraphics[height=45mm]{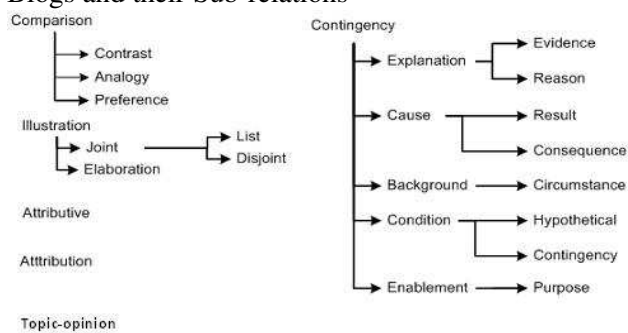}
\end{figure}
%\vspace{-6mm}

\paragraph{Illustration:} Is used to provide additional information or detail
    about a situation. For example:      \emph{``Allied Capital is a closed-end management
investment company that will operate as a business development concern.''}

As shown in Figure~\ref{rr},
\emph{illustration} relations can be sub-divided into sub-categories: \emph{joint}, \emph{list}, \emph{disjoint}, and
\emph{elaboration} relations according to the RST Discourse Treebank \cite{Carlson01} and the Penn Discourse TreeBank \cite{Prasad08}.

\paragraph{Contingency:} Provides cause, condition, reason or evidence for a
situation, result or claim. For example: \emph{``The meat is good because
they slice it right in front of you.''}

As shown in Figure~\ref{rr},
the \emph{contingency}
relation subsumes several more specific relations: \emph{explanation}, \emph{evidence},
\emph{reason}, \emph{cause}, \emph{result}, \emph{consequence},
\emph{background}, \emph{condition}, \emph{hypothetical},
\emph{enablement}, and \emph{purpose} relations according to the Penn Discourse TreeBank \cite{Prasad08}.

\paragraph{Comparison:} Gives a comparison
and contrast among different situations. For example, \emph{``Its fast-forward and rewind work much more smoothly and consistently than those of other models I've had.''}

The \emph{comparison} relation
subsumes the \emph{contrast} relation according to the Penn Discourse TreeBank~\cite{Prasad08} and the \emph{analogy} and \emph{preference} relations according to the RST Discourse Treebank~\cite{Carlson01}.

\paragraph{Attributive:} Relation provides details about an entity or an event - e.g. \emph{``Mary has a pink coat.''}. It can be
used to illustrate a particular feature about a concept or an entity - e.g. \emph{``Picasa makes sure your pictures are always organized.''}. The \emph{attributive} relation, also included in Grimes'
predicates~\cite{Grimes75}, is considered because it describes
attributes or features of an object or event and is often used in
query-based summarization and question answering. 

\paragraph{Topic-opinion:} We introduced
topic-opinion relations to represent opinions which are not expressed by reported speech. This relation can be used to express an opinion: an internal feeling or belief towards an object or an
event. For example: \emph{``Cage is a wonderfully versatile actor.''}

\paragraph{Attribution:} These relations are instances of reported speech both direct and indirect which may express feelings, thoughts, or hopes. For example: \emph{``The legendary GM chairman declared that his company would make ``a car for every purse and purpose.''''}

\subsection{Automatic Discourse Tagging}
Once the manual analysis identified the most prevalent set of relations, we tried to measure their frequency by tagging them automatically within a larger corpus. Only recently, the HILDA~\cite{hilda} and~\cite{feng12}'s discourse parser were made publicly available. Both of these parsers work at the text-level, as opposed to the sentence-level, and hence currently achieve the highest tagging performance when compared to the state of the art. \cite{feng12}'s work showed a significant improvement on the performance of HILDA by enhancing its original feature set. However, at the time this research was done, the only publicly available discourse
parser was SPADE~\cite{Soricut03} which operates on individual sentences. To identify \emph{illustration}, \emph{contingency}, \emph{comparison}, and \emph{attribution} relations, we have used SPADE discourse parser.  However, we have complemented this parser with three other approaches: \cite{Jindal06}'s approach is used to identify intra-sentence
\emph{comparison} relations; we have designed a tagger based on~\cite{Fei06}'s approach to
identify \emph{topic-opinion} relations; and we have proposed a new
approach to tag \emph{attributive} relations~\cite{Mit12}. A description and evaluation of
these approaches can be found in~\cite{Mit12}. By combining these
approaches, a sentence is tagged with all possible discourse relations that it contains. 

\subsection{Distribution of Discourse Relations}
To find the most prevalent discourse relations for opinion summarization, we have used the TAC 2008 opinion summarization track input document set (collection) which is a subset of BLOG06 and the answer nuggets provided by TAC 2008 as the reference summary (or model summaries), which had been created
to evaluate participants' summaries at the TAC 2008 opinion summarization track. The collection consists of 600 blogs on 28 different topics. The dataset of the model summaries consists of 693 sentences.

Using the discourse parsers presented in Section 3.2, we computed the distribution of discourse relations within the TAC 2008 opinion summarization collection and the model summaries. \emph{Illustration}, \emph{contingency}, \emph{comparison}, \emph{attributive}, \emph{topic-opinion}, and \emph{attribution} are the most frequently occuring relations in our data sets. The distribution is shown in Table~\ref{disD} \footnote{In Table \ref{disD}, the percentages do not add up to 100 because a sentence may contain more than one relation.}. 

%\vspace{-2mm}
\begin{table}[h]
\begin{center}
\begin{small}
\caption{Distribution of Discourse Relations in the TAC-2008 and DUC-2007 Datasets}\label{disD}
%\vspace{-3mm}
\begin{tabular}[t]{|l|r|r|r|r|}
\hline
\textbf{Discourse}&\multicolumn{2}{|c|}{\textbf{TAC 2008}} &
\multicolumn{2}{|c|}{\textbf{DUC 2007}}\\
% \hline
% &\textbf{TAC 2008}& \textbf{TAC 2008} &\textbf{DUC 2007}& \textbf{DUC 2007}\\
\textbf{Relation}& \textbf{Coll.} & \textbf{Model} & \textbf{Coll.} & \textbf{Model}\\
 \hline
Illustration &52\%& 46\% &42\%&38\%\\
Contingency&31\% & 37\%&34\% &29\%\\
Comparison&23\%&18\%&15\% &12\%\\
Attributive &12\%& 28\%&3\%  &4\%\\
Topic-opinion&14\%&15\%&4\%&5\%\\
Attribution&11\%&9\%&2\%&3\%\\ \hline
other&13\%&9\%&28\%&31\%\\
none&14\%&10\%&8\%&7\%\\ \hline
\end{tabular}
\end{small}
\end{center}
\end{table}
%\vspace{-2mm}

Table~\ref{disD} shows that in the TAC 2008 input document set, the \emph{illustration} relation occurs in 52\% of the sentences; while \emph{attribution} is the least frequently occurring relation. In this dataset, other relations, such as \emph{antithesis} and \emph{temporal} relations, occur in about 13\% of the sentences and about 14\% of the sentences did not receive any relation tag. As indicated in Table~\ref{disD}, the TAC model summaries have a similar distribution as the collection as a whole. The \emph{attributive} relation seems, however, to be more frequent in the summaries (28\%) than in the original texts (12\%). We suspect that the reason for this is due to the question types of this track. To successfully generate query-relevant summaries that answer the questions of this track, candidate sentences need to contain \emph{attributive} relations. For example, to answer the questions from this track \emph{``Why do people like Picasa?''} or \emph{``What features do people like about Windows Vista?''}, the summary needs to provide details about these entities or illustrate a particular feature about them. As a result, the summary will be composed of many \emph{attributive} relations since \emph{attributive} relations help to model the required information.

To compare the distribution of discourse relations within more formal types of texts such as news articles, we used
the Document Understanding Conference (DUC) 2007 Main Task input document set (collection) and their associated model summaries. The DUC 2007 dataset is a news article based dataset from the AQUAINT corpus. The DUC 2007 input document set contains 1125 news articles on 45 different topics. The model summaries were used to evaluate the DUC 2007 participants' summaries. The dataset of the model summaries contains 180 summaries generated by the National Institute of Standards and Technology (NIST) assessors with a summary length of about 250 words. The distribution of relations in this dataset are shown in Table~\ref{disD}.

Table~\ref{disD} shows that the most frequently occurring relation in the DUC 2007 document collection and in the model summaries is \emph{illustration}; while the \emph{attribution} relation is the least frequently occurring relation. Here again, it is interesting to note that the distribution of the discourse relations in the document collection and in the model summaries is generally comparable.

The distribution of the \emph{illustration}, \emph{contingency}, and \emph{comparison} relations in the DUC 2007 dataset is comparable to those in the TAC 2008 opinion summarization dataset. Indeed, Table~\ref{disD} shows that \emph{illustration}, \emph{contingency}, and \emph{comparison} relations occur quite frequently irrespective of the textual genre. However, in contrast to the TAC dataset, \emph{attributive}, \emph{topic-opinion}, and \emph{attribution} relations occur very rarely in DUC 2007. We suspect that this is mostly due to the opinionated nature of blogs. Another observation is that \emph{temporal} relations (included in ``other'') occurred very frequently (30\%) in the DUC 2007 dataset whereas this relation occurs rarely in the blog dataset. This is in line with our intuition that news articles present events that inherently contain temporal information.

\section{Evaluation of Discourse Relations}
To measure the usefulness of discourse relations for the summarization of informal texts, we have tested the effect of each relation with four different summarizers: BlogSum~\cite{Mit12}, MEAD~\cite{Radev04}, the best scoring system at TAC 2008\footnote{http://www.nist.gov/tac/} and the best scoring system at DUC 2007\footnote{http://www-nlpir.nist.gov/projects/duc/guidelines/2007.html}. We have evaluated the effect of each discourse relation on the summaries generated and compared the results. Let us first describe the BlogSum summarizer.

\subsection{BlogSum}
BlogSum is a domain-independent query-based extractive summarization  system that uses intra-sentential discourse relations within the framework based on text schemata. The heart of BlogSum is based on discourse relations and text schemata.

%Text schemata  are patterns of discourse organization used to achieve different communicative goals. Text schemata were first introduced by McKeown~\cite{McKeown85} based on the observation that specific types of discourse schemata are more effective to achieve a particular communicative goal. Schema-based approaches were also used by other researchers in the context of question answering and text generation to generate relevant and coherent text \cite{Cline94}. However, schema-based approaches are usually
%domain-dependent where the domain knowledge is pre-compiled and explicitly represented in knowledge bases or is used for structured documents (e.g. Wikipedia articles).
BlogSum works in the following way: First candidate sentences are extracted and ranked using the
topic and question similarity to give priority to topic and question relevant sentences. Since BlogSum has been designed for blogs, which are opinionated in nature, to rank a sentence, the sentence polarity (e.g. positive, negative or neutral) is calculated and used for sentence ranking. To extract and rank sentences, BlogSum thus calculates a
score for each sentence using the features shown below:\\

\vspace{-3mm}
\noindent\emph{Sentence Score = }\noindent\emph{$w_1$ $\times$ Question Similarity + \\
\hspace*{2.8cm}$w_2$ $\times$ Topic Similarity + \\
\hspace*{2.8cm}$w_3$ $\times$ Subjectivity Score}\\

\vspace{-3mm}
where, question similarity and topic similarity are calculated using the cosine similarity based on words \emph{tf.idf} and the subjectivity score is calculated using a dictionary-based approach based on the MPQA lexicon\footnote{MPQA:
http://www.cs.pitt.edu/mpqa}.  Once sentences are ranked, they are categorized based
on the discourse relations that they convey. This step is critical because the automatic identification of discourse relations renders BlogSum independent of the domain. This step also plays a key role in content selection and summary coherence as schemata are designed using these relations.

\begin{figure}[h]
\centering
\caption{A Sample Discourse Schema used in BlogSum}
\vspace{-2mm}
~\\
\label{sch}
\includegraphics[height=36mm]{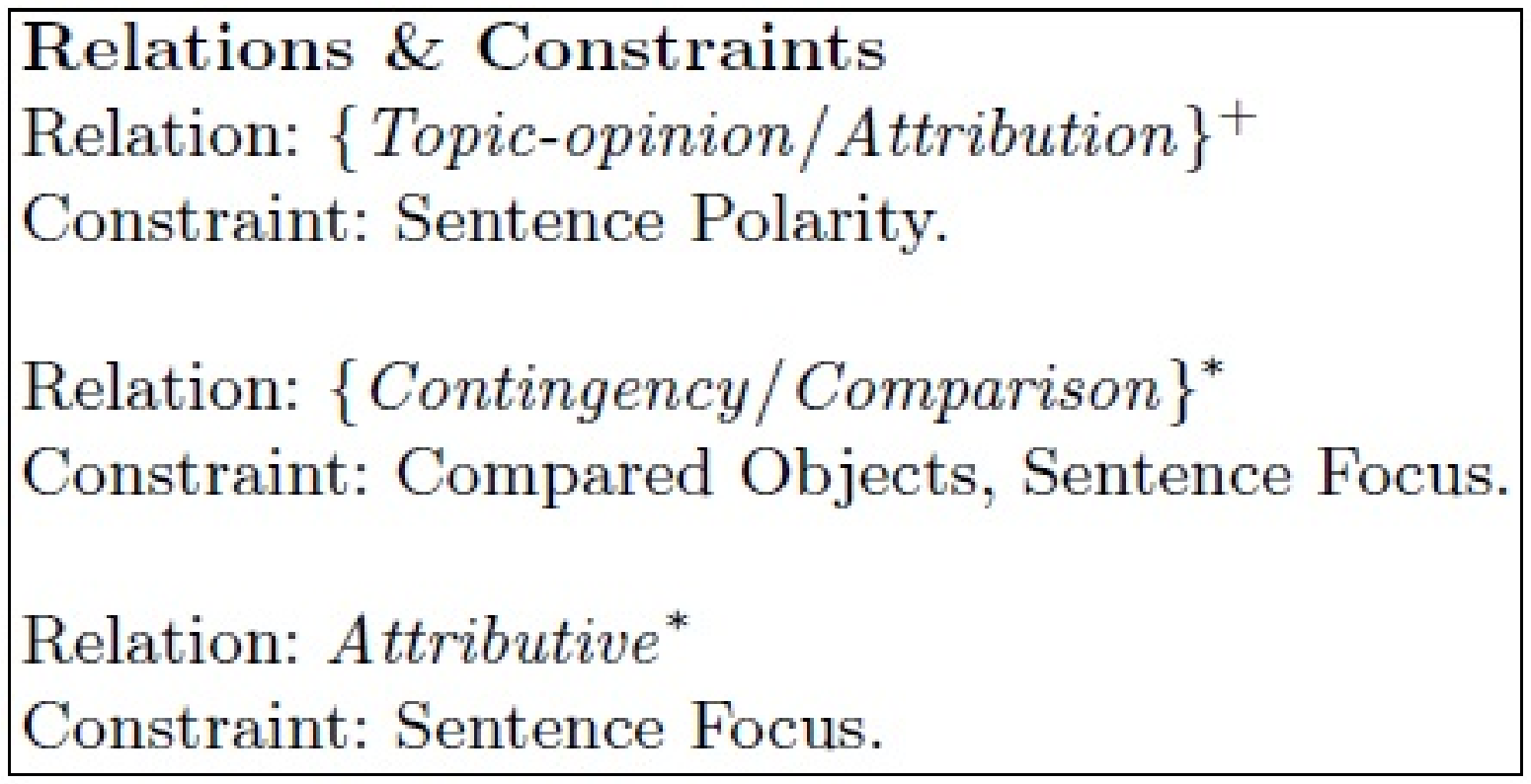}
 \end{figure}

In order not to answer all questions the same way, BlogSum uses different schemata to generate a summary that answers specific types
of questions. Each schema is designed to give priority to its associated question type and subjective
sentences as summaries for opinionated texts are generated. Each schema specifies the types of discourse relations and the order in which they should appear in
the output summary for a particular question type. Figure~\ref{sch} shows a sample schema that is used to answer \emph{reason} questions (e.g. ``\emph{Why do people like Picasa?}''). According to this schema\footnote{The notation / indicates an alternative, \{ \} indicates optionality, * indicates that the item may appear 0 to n times and + indicates that the item may appear 1 to n times}, one or more sentences containing a \emph{topic-opinion} or \emph{attribution} relation
followed by zero or many sentences containing a \emph{contingency} or
\emph{comparison} relation followed by zero or many sentences containing a
\emph{attributive} relation should be used.

\vspace{-3mm}
\vspace{-3mm}

Finally the most appropriate schema is selected based on a given question type; and candidate sentences fill particular slots in the selected schema based on which discourse relations they contain in order to create the final summary (details of BlogSum can be found in~\cite{Mit12}).

\subsection{Evaluation of Discourse Relations on Blogs}
To evaluate the effect of each discourse relation for blog summarization, we performed several experiments. We used as a baseline the original ranked list of candidate
sentences produced by BlogSum before applying the discourse schemata, and compared this to the BlogSum-generated summaries with and without each discourse relation. We used the TAC 2008 opinion summarization dataset which consists of 50 questions on 28
topics; on each topic one or two questions were asked and 9 to 39
relevant documents were given. For each question, one summary was
generated with no regards to discourse relations and two summaries were produced by BlogSum: one using the discourse tagger and the other without using the specific discourse tagger. The maximum summary
length was restricted to 250 words.

To measure the effect of each relation, we have automatically evaluated how BlogSum performs using the standard ROUGE-2 and ROUGE-SU4 measures. For comparative purposes, Table~\ref{tac} shows the official ROUGE-2 (R-2) and ROUGE-SU4 (R-SU4) for all 36 submissions of the TAC 2008 opinion summarization track.  In the table, ``TAC Average'' refers to the mean performance of all participant systems and ``TAC-Best'' refers to the best-scoring system at TAC 2008.

\vspace{-3mm}
\begin{table}[h]
\caption{Results  of the TAC 2008 Opinion Summarization Track} \label{tac}
\vspace{-4mm}
\begin{center}
\begin{small}
\begin{tabular}[t]{|l|c|c|}
\hline
\textbf{System Name} &\textbf{R-2} & \textbf{R-SU4}\\
 \hline \hline
TAC Average & 0.069&0.086 \\
TAC-Best&0.130&0.139\\
 \hline
\end{tabular}
\end{small}
\end{center}
\end{table}
\vspace{-3mm}
%
%\vspace{-3mm}
%\begin{table}[h]
%\caption{Effect of Discourse Relations on Summarization on the TAC 2008 Dataset} \label{evalContent}
%\vspace{-6mm}
%\begin{center}
%\begin{tiny}
%\begin{tabular}[t]{|l||l|l||l|l||l|l|}
%\hline
%\textbf{System Name} &\multicolumn{2}{c||} {\bf BlogSum} &\multicolumn{2}{c||} {\bf MEAD} &\multicolumn{2}{c|} {\bf TAC-Best}  \\
% &\textbf{R-2} & \textbf{R-SU4}  &\textbf{R-2} & \textbf{R-SU4}  &\textbf{R-2} & \textbf{R-SU4}\\
% \hline \hline
% Baseline& .102 & .107  &0.041&0.064&.130&.139\\ \hline
% \hline
% w/o Illustration  & .107$\Downarrow$ & .110$\Downarrow$ &0.022$\Downarrow$&0.041$\Downarrow$&.112$\Downarrow$&.120$\Downarrow$\\
%\hline
% w/o Contingency  & .093$\Downarrow$ & .102$\Downarrow$ &0.025$\Downarrow$&0.046$\Downarrow$&.102$\Downarrow$&.110$\Downarrow$\\
%\hline
% w/o Comparison  & .103$\Downarrow$ & .108$\Downarrow$ &0.033$\Downarrow$&0.052$\Downarrow$&.113$\Downarrow$&.122$\Downarrow$\\
%\hline
% w/o Attributive  & .113$\Downarrow$ & .115$\Downarrow$ &0.050&0.072&.124&.130\\
%\hline
% w/o Topic-opinion  & .112$\Downarrow$ & .117 &0.049&0.072&.123&.129\\
%\hline
% w/o Attribution  & .118$\Downarrow$  & .127$\Downarrow$ &0.051$\Downarrow$&0.073$\Downarrow$&.128&.132\\
%\hline
%\hline
% with all Relations & \textbf{.125} & \textbf{.128} & \textbf{.053}&\textbf{.075}&\textbf{.138}&\textbf{.151}\\
%\hline
%\end{tabular}
%\end{tiny}
%\end{center}
%\end{table}
%\vspace{-2mm}

\vspace{-3mm}
\begin{table}[h]
\caption{Effect of Discourse Relations on ROUGE-2 with the TAC 2008 Dataset} \label{evalContentR2}
\vspace{-3mm}
\begin{center}
\begin{small}
\begin{tabular}[t]{|l|l|l|l|}
\hline
\textbf{System Name} & {\bf BlogSum} & {\bf MEAD} & {\bf TAC-Best}  \\
 &\textbf{R-2}  &\textbf{R-2} &  \textbf{R-2} \\
 \hline \hline
 Baseline& 0.102$\Downarrow$ & 0.041$\Downarrow$&0.130\\ \hline
 \hline
 w/o Illustration  & 0.107$\Downarrow$ & 0.022$\Downarrow$&0.112$\Downarrow$\\
\hline
 w/o Contingency  & 0.093$\Downarrow$ & 0.025$\Downarrow$&0.102$\Downarrow$\\
\hline
 w/o Comparison  & 0.103$\Downarrow$ & 0.033$\Downarrow$&0.113$\Downarrow$\\
\hline
 w/o Attributive  & 0.113$\Downarrow$ &0.050&0.124\\
\hline
 w/o Topic-opinion  & 0.112$\Downarrow$ & 0.049&0.123\\
\hline
 w/o Attribution  & 0.118$\Downarrow$  &0.051$\Downarrow$&0.128\\
\hline
\hline
 with all Relations & \textbf{0.125} &  \textbf{.053}&\textbf{0.138}\\
\hline
\end{tabular}
\end{small}
\end{center}
\end{table}
\vspace{-2mm}

\vspace{-3mm}
\begin{table}[h]
\caption{Effect of Discourse Relations on ROUGE-SU4 with the TAC 2008 Dataset} \label{evalContentRSU4}
\vspace{-2mm}
\begin{center}
\begin{small}
\begin{tabular}[t]{|l|l|l|l|}
\hline
\textbf{System Name} &{\bf BlogSum} &{\bf MEAD} &{\bf TAC-Best}  \\
 & \textbf{R-SU4}  & \textbf{R-SU4}  & \textbf{R-SU4}\\
 \hline \hline
 Baseline& 0.107$\Downarrow$  &0.064$\Downarrow$&0.139\\ \hline
 \hline
 w/o Illustration  & 0.110$\Downarrow$ &0.041$\Downarrow$&0.120$\Downarrow$\\
\hline
 w/o Contingency  & 0.102$\Downarrow$ &0.046$\Downarrow$&0.110$\Downarrow$\\
\hline
 w/o Comparison  &  0.108$\Downarrow$ &0.052$\Downarrow$&0.122$\Downarrow$\\
\hline
 w/o Attributive  &  0.115$\Downarrow$ &0.072&0.130\\
\hline
 w/o Topic-opinion  & 0.117 &0.072&0.129\\
\hline
 w/o Attribution  &  0.127$\Downarrow$ &0.073$\Downarrow$&0.132\\
\hline
\hline
 with all Relations & \textbf{0.128} & \textbf{0.075}&\textbf{0.151}\\
\hline
\end{tabular}
\end{small}
\end{center}
\end{table}
\vspace{-2mm}

The results of our evaluation are shown in Tables~\ref{evalContentR2} (ROUGE-2) and~\ref{evalContentRSU4} (ROUGE-SU4).  As the tables show, BlogSum's baseline is situated below the best scoring system at TAC-2008, but much higher than the average system (see~Table~\ref{tac}); hence, it represents a fair baseline.  The tables further show that using both the ROUGE-2 (R-2) and ROUGE-SU4 (R-SU4) metrics,  with the TAC 2008 dataset, BlogSum performs better when taking discourse relations into account. Indeed, when ignoring discourse relations, BlogSum has a R2=0.102 and R-SU4=0.107 and misses many question relevant sentences; whereas the inclusion of these relations helps to incorporate those relevant sentences into the final summary and brings the R-2 score to 0.125 and R-SU4 to 0.128. In order to verify if these improvements were statistically significant, we performed a 2-tailed t-test.  The results of this test are indicated with the $\Downarrow$ symbol in Tables~\ref{evalContentR2} and ~\ref{evalContentRSU4}.  For example, the baseline setup of BlogSum performed significantly lower for both R-2 and R-SU4 compared to BlogSum with all relations. This result indicates that the use of discourse relations as a whole helps to include more question relevant sentences and improve the summary content.

To ensure that the results were not specific to our summarizer, we performed the same experiments with two other systems: the MEAD summarizer \cite{Radev04}, a publicly available and a widely used summarizer, and with the output of the TAC best-scoring system. For MEAD, we first generated candidate sentences using MEAD, then these candidate sentences were tagged using discourse relation taggers used under BlogSum. Then these tagged sentences were filtered using BlogSum so that no sentence with a specific relation is used in summary generation for a particular experiment. We have calculated ROUGE scores using the original candidate sentences generated by MEAD and also using the filtered candidate sentences. As a baseline, we used the original candidate sentences generated by MEAD. As a best case scenario, we have passed these candidate sentences through the discourse schemata used by BlogSum (see Section 4.1). In Tables~\ref{evalContentR2} and~\ref{evalContentRSU4}, this is referred to as ``MEAD with all relations''. We have applied the same approach with the output of the TAC best-scoring system. In the tables, ``TAC-Best Baseline'' refers to the original summaries generated by the TAC-Best system and ``TAC-Best with all relations'' refers to the summaries generated by applying discourse schemata using the summary sentences generated by the TAC-Best system.

When looking at individual relations, Tables~\ref{evalContentR2} and~\ref{evalContentRSU4} show that considering \emph{illustrations}, \emph{contingencies} and  \emph{comparisons}  make a statistically significant improvement in all scenarios, and with all summarisers.   For example, if TAC-Best does not consider \emph{illustration} relations, then the R-2 score decreases from 0.138 to 0.112, 0.102 and 0.113, respectively. On the other hand, the relations of \emph{topic-opinion}, \emph{attribution}, and \emph{attributive} do not consistently lead to a statistically significant improvement on ROUGE scores.

It is interesting to note that although informal texts may not exhibit a clear discourse structure, the use of individual discourse relations such as \emph{illustration}, \emph{contingency} and \emph{comparison} is nonetheless useful in the analysis of informal documents such as those found in the social media.

\subsection{Effect of Discourse Relations on News}
To compare the results found with blogs with more formal types of texts, we have performed the same experiments but, this time with the DUC 2007 Main Task dataset. In this task, given a topic (title) and a set of 25 relevant documents, participants had to create an automatic summary of length 250 words from the input documents. In the dataset, there were 45 topics and thirty teams participated to this shared task. Table~\ref{duc} shows the official ROUGE-2 (R-2) and ROUGE-SU4 (R-SU4) scores of the DUC 2007 main task summarization track. In Table~\ref{duc}, ``DUC Average'' refers to the mean performance of all participant systems and ``DUC-Best'' refers to the best scoring system at DUC 2007. 

%\vspace{-3mm}
\begin{table}[h]
\caption{DUC 2007 Main Task Summarization Results} \label{duc}
\vspace{-4mm}
\begin{center}
\begin{small}
\begin{tabular}[t]{|l|c|c|}
\hline
\textbf{System Name} &\textbf{R-2} & \textbf{R-SU4}\\
 \hline \hline
DUC Average & 0.095&0.157\\
 \hline
DUC-Best & 0.124&0.177 \\
 \hline
\end{tabular}
\end{small}
\end{center}
\end{table}
%\vspace{-3mm}

%%\vspace{-3mm}
%\begin{table}[h]
%\caption{Effect of Discourse Relations on Summarization on the DUC 2007 Dataset} \label{evalDUC}
%\vspace{-6mm}
%\begin{center}
%\begin{tiny}
%\begin{tabular}[t]{|l||l|l||l|l||l|l|}
%\hline
%\textbf{System Name} &\multicolumn{2}{c||} {\bf BlogSum} &\multicolumn{2}{c||} {\bf MEAD} &\multicolumn{2}{c|} {\bf DUC-Best}  \\
% &\textbf{R-2} & \textbf{R-SU4}  &\textbf{R-2} & \textbf{R-SU4}  &\textbf{R-2} & \textbf{R-SU4}\\
% \hline \hline
%Baseline & .089 & .110 &0.099&.142&.125&.177\\
%\hline \hline
%w/o Illustration  & .079$\Downarrow$ & .117$\Downarrow$ & .061$\Downarrow$&.118$\Downarrow$&.103$\Downarrow$ & .138$\Downarrow$\\
%\hline
%w/o Contingency  & .074$\Downarrow$ & .113$\Downarrow$ &0.060$\Downarrow$&.118$\Downarrow$&0.097$\Downarrow$ & .123$\Downarrow$\\
%\hline
%w/o Comparison  & .086$\Downarrow$ & .122$\Downarrow$ &0.078$\Downarrow$&.130$\Downarrow$&.114$\Downarrow$ &  .144$\Downarrow$\\
%\hline
%w/o Attributive  & .092 & .131 &0.099&.141$\Downarrow$&.119$\Downarrow$ & .159$\Downarrow$\\
%\hline
%w/o Topic-opinion  & .092 & .130 &0.099&.141$\Downarrow$&.115$\Downarrow$ & .153$\Downarrow$\\
%\hline
%w/o Attribution  & .093 & .131 &0.099&.142$\Downarrow$&.120$\Downarrow$ &.164$\Downarrow$\\
%\hline
%\hline
%with all Relations & \textbf{.093} & \textbf{.132} &\textbf{.110}&\textbf{.168}&\textbf{.157}& \textbf{.196}\\
%\hline
%\end{tabular}
%\end{tiny}
%\end{center}
%\end{table}

\begin{table}[h]
\caption{Effect of Discourse Relations on ROUGE-2 with the DUC 2007 Dataset} \label{evalDUCR2}
%\vspace{-6mm}
\begin{center}
\begin{small}
\begin{tabular}[t]{|l||l||l||l|}
\hline
\textbf{System Name} & {\bf BlogSum} & {\bf MEAD} & {\bf DUC-Best}  \\
 &\textbf{R-2}  &\textbf{R-2}   &\textbf{R-2} \\
 \hline \hline
Baseline & 0.089 &0.099&0.124$\Downarrow$\\
\hline \hline
w/o Illustration  & 0.079$\Downarrow$  & 0.061$\Downarrow$&0.103$\Downarrow$ \\
\hline
w/o Contingency  & 0.074$\Downarrow$ &0.060$\Downarrow$&0.097$\Downarrow$ \\
\hline
w/o Comparison  & 0.086$\Downarrow$ &0.078$\Downarrow$&0.114$\Downarrow$\\
\hline
w/o Attributive  & 0.092  &0.099&0.119$\Downarrow$ \\
\hline
w/o Topic-opinion  & 0.092  &0.099&0.115$\Downarrow$ \\
\hline
w/o Attribution  & 0.093  &0.099&0.120$\Downarrow$ \\
\hline
\hline
with all Relations & \textbf{0.093}  &\textbf{0.110}&\textbf{0.157}\\
\hline
\end{tabular}
\end{small}
\end{center}
\end{table}

\begin{table}[h]
\caption{Effect of Discourse Relations on ROUGE SU-4 with the DUC 2007 Dataset} \label{evalDUCRSU4}
\vspace{-6mm}
\begin{center}
\begin{small}
\begin{tabular}[t]{|l|l|l|l|}
\hline
\textbf{System Name} & {\bf BlogSum} & {\bf MEAD} & {\bf DUC-Best}  \\
 & \textbf{R-SU4}  & \textbf{R-SU4}  & \textbf{R-SU4}\\
 \hline \hline
Baseline &  0.110$\Downarrow$ &0.142$\Downarrow$&0.177$\Downarrow$\\
\hline \hline
w/o Illustration  &  0.117$\Downarrow$ &0.118$\Downarrow$ & 0.138$\Downarrow$\\
\hline
w/o Contingency  & 0.113$\Downarrow$ &0.118$\Downarrow$& 0.123$\Downarrow$\\
\hline
w/o Comparison  &  0.122$\Downarrow$ &0.130$\Downarrow$&  0.144$\Downarrow$\\
\hline
w/o Attributive  &  0.131 &0.141$\Downarrow$& 0.159$\Downarrow$\\
\hline
w/o Topic-opinion   & 0.130 &0.141$\Downarrow$& 0.153$\Downarrow$\\
\hline
w/o Attribution  & 0.131 &0.142$\Downarrow$&0.164$\Downarrow$\\
\hline
\hline
with all Relations & \textbf{0.132} &\textbf{0.168}& \textbf{0.196}\\
\hline
\end{tabular}
\end{small}
\end{center}
\end{table}
%\vspace{-3mm}

Tables~\ref{evalDUCR2} and~\ref{evalDUCRSU4}  show the results with this dataset with respect to ROUGE-2 and ROUGE-SU4, respectively. As the tables show, BlogSum's performance with all discourse relations (R2=0.093 and R-SU4=0.132) is similar to the DUC average performance shown in Table~\ref{duc} (R2=00.095 and R-SU4=0.157) which is much lower than the DUC-Best performance (R2=0.124, R-SU4=0.177) shown in Table~\ref{duc}). However, these results show that even though BlogSum was designed for informal texts, it still performs relatively well with formal documents. Tables~\ref{evalDUCR2} and ~\ref{evalDUCRSU4} further show that with the news dataset, the same relations have the most effect as with blogs. Indeed BlogSum generated summaries also benefit most from the \emph{contingency}, \emph{illustration}, and \emph{comparison} relations; and all three  relations bring a statistically significant contribution to the summary content. 

Here again, as shown in Tables~\ref{evalDUCR2} and ~\ref{evalDUCRSU4}, we performed the same experiments with two other systems: the MEAD summarizer and the output of the DUC-Best system.  Again, for the DUC 2007 dataset, each discourse relation has the same effect on summarization with all systems as with the blog dataset: \emph{contingency}, \emph{illustration}, and \emph{comparison} provide a statistically significant improvement in content; while \emph{attributive}, \emph{topic-opinion} and \emph{attribution} do not reduce the content, but do not see to bring a systematic and significant improvement. \\

\section{Conclusion and Future Work}
In this paper, we have evaluated the effect of discourse relations on summarization. We have considered the six most frequent relations in blogs - namely \emph{comparison}, \emph{contingency}, \emph{illustration}, \emph{attribution}, \emph{topic-opinion}, and \emph{attributive}. First, we have measured the distribution of discourse relations on blogs and on news articles and show that the prevalence of these six relations is not genre dependent. For example, the relations of \emph{illustration}, \emph{contingency}, and \emph{comparison} occur frequently in both textual genres. We have then evaluated the effect of these six relations on summarization with the TAC 2008 opinion summarization dataset and the DUC 2007 dataset. We have conducted these evaluations with our summarization system called BlogSum, the TAC best-scoring system, the DUC best-scoring system, and the MEAD summarizer. The results show that for both textual genres, some relations have more effect on summarization compared to others. In both types of texts, the \emph{contingency}, \emph{illustration}, and \emph{comparison} relations provide a significant improvement on summary content; while the \emph{attribution}, \emph{topic-opinion}, and \emph{attributive} relations do not provide a systematic and statistically significant improvement. These results seem to indicate that, at least for summarization, discourse relations are just as useful for informal and affective texts as for more traditional news articles. This is interesting, because although informal texts may not exhibit a clear discourse structure, the use of individual discourse relations is nonetheless useful in the analysis of informal documents. 

In the future, it would be interesting to evaluate the effect of other relations such as the \emph{temporal} relation. Indeed, \emph{temporal} relations occur infrequently in blogs but are very frequent in news articles. Such an analysis would allow us to tailor the type of discourse relations to include in the final summary as a function of the textual genre being considered. In the future, it would also be interesting to use other types of texts such as reviews and evaluate the effect of discourse relations using other measures than ROUGE-2 and ROUGE-SU4. Finally, we would like to validate this work again with the newly available discourse parsers of~\cite{hilda} and~\cite{feng12}.

\section*{Acknowledgement}
The authors would like to thank the anonymous referees for their valuable comments on an earlier version of the paper. This work was financially supported by an NSERC grant.

%Another interesting research avenue is the analysis of inter-sentential discourse relations. Indeed, the present paper %focused only on intra-sentential relations. Relations across sentences have the potential to increase summary coherence %and cohesion further, and may also have a positive effect on summary content.

\end{document}